\documentclass[10pt,journal,compsoc]{IEEEtran}
\IEEEoverridecommandlockouts

\usepackage[noadjust]{cite}
\usepackage[T1]{fontenc}
\usepackage{amsmath}
\usepackage{amssymb}
\usepackage{array}
\usepackage{booktabs}
\usepackage[table]{xcolor}
\usepackage{graphicx}
\usepackage{microtype}
\usepackage{tabularx}
\usepackage{flushend}
\usepackage[ruled,vlined]{algorithm2e}
\usepackage{xspace}
\usepackage[
  bookmarks=true,
  bookmarksnumbered=true,
  pdfpagemode=UseOutlines,
  plainpages=false,
  pdfpagelabels=true,
  hidelinks
]{hyperref}

\microtypesetup{protrusion=true,expansion=true,kerning=true,spacing=true}
\makeatletter
\def\@IEEEBIOskipN{8\baselineskip}
\makeatother

\newcommand{\SPDL}{\emph{SPD matrix learning}\xspace}

\hypersetup{
  pdftitle={SPD Matrix Learning for Neuroimaging Analysis: Perspectives, Methods, and Challenges},
  pdfsubject={Survey of symmetric positive definite matrix learning methods for neuroimaging analysis},
  pdfauthor={Ce Ju, Reinmar Kobler, Antoine Collas, Motoaki Kawanabe, Cuntai Guan, and Bertrand Thirion},
  pdfkeywords={Symmetric positive-definite matrices, geometric deep learning, geometric statistics, Riemannian geometry, brain-computer interfaces, neuroimaging, disease progression modeling, longitudinal modeling}
}

\begin{document}

\title{SPD Matrix Learning for Neuroimaging Analysis: \\
Perspectives, Methods, and Challenges
}

\author{Ce~Ju$^{\dagger}$,
Reinmar~Kobler$^{\ddagger}$,
Antoine~Collas$^{\dagger}$,\\
Motoaki~Kawanabe$^{\ddagger}$,
Cuntai~Guan$^{\S}$,
and Bertrand~Thirion$^{\dagger}$
\thanks{$^{\dagger}$\,Inria, CEA, Université Paris-Saclay, Palaiseau, France. Emails: \{ce.ju, antoine.collas, bertrand.thirion\}@inria.fr.}
\thanks{$^{\ddagger}$\,Advanced Telecommunications Research Institute International and RIKEN Artificial Intelligence Project, Japan. Emails: \{reinmar.kobler,kawanabe\}@atr.jp.}
\thanks{$^{\S}$\,College of Computing and Data Science and Centre of AI in Medicine, Nanyang Technological University, Singapore. Email: ctguan@ntu.edu.sg.}
}

\maketitle

\IEEEdisplaynontitleabstractindextext

\begin{abstract}
Neuroimaging provides essential tools for characterizing brain activity, structure, and connectivity through modalities that capture complementary aspects of brain organization. Across these diverse modalities, a unifying perspective arises when measurements are modeled as symmetric positive-definite (SPD)-valued representations through appropriate estimation or regularization procedures. Endowed with Riemannian geometry, the SPD manifold provides a non-Euclidean framework for principled statistical inference and machine learning on these representations. This review organizes these analytical and learning approaches within a framework for SPD matrix learning that connects classical geometric statistics with modern machine learning across neuroimaging and neurophysiological applications. We systematically survey the progression from modality-specific representations to geometric shallow and deep learning paradigms, highlighting how SPD matrix learning preserves underlying structural constraints while extending to modern AI applications in neuroimaging and brain-computer interfaces.

\end{abstract}


\section{INTRODUCTION}

 \begin{figure*}
   \centering
   \includegraphics[width=0.866\textwidth]{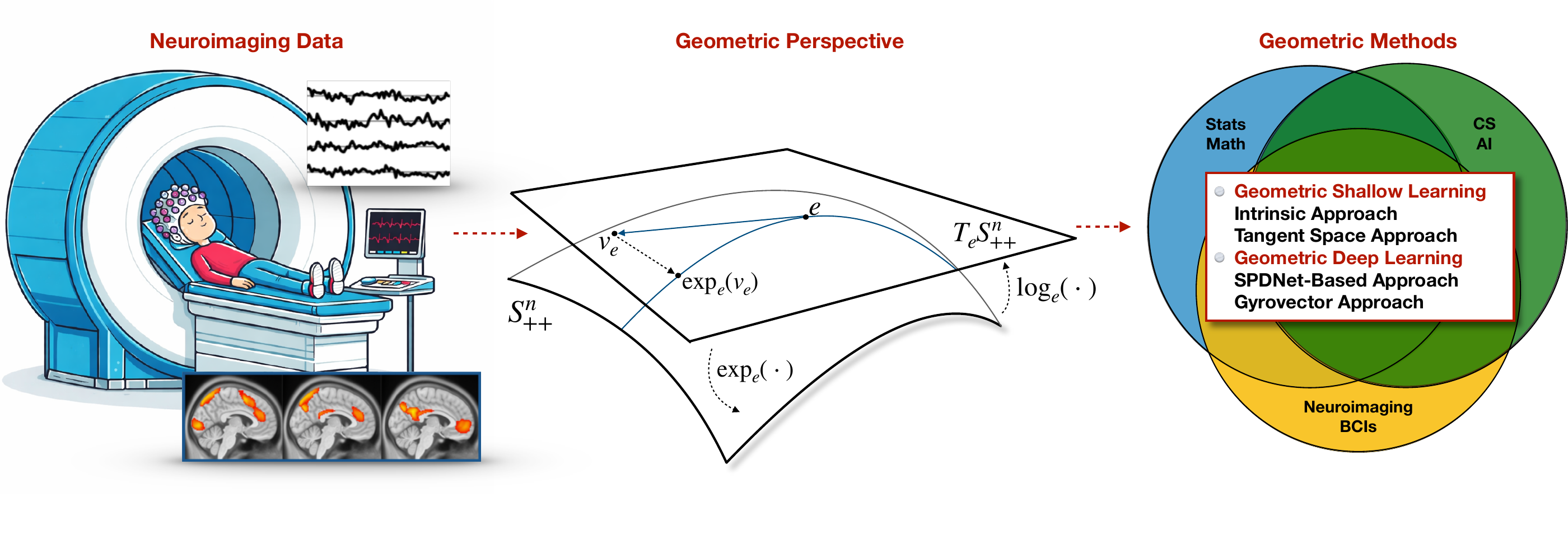}
   \caption{SPD Matrix Learning Paradigm. Neuroimaging measurements from various modalities (e.g., M/EEG, ECoG, fMRI, MRI, and DTI) can be modeled as points on the SPD manifold $\mathcal{S}_{++}^n$ through appropriate estimation or regularization. The framework illustrates the manifold geometry, its tangent space at a Fr\'echet mean $e$, a tangent vector, a geodesic, and the exponential and logarithmic maps connecting the manifold with its tangent space. Within this framework, SPD matrix learning bridges classical geometric statistics with modern machine learning to analyze SPD-valued neuroimaging data.}
   \label{fig:spd_learning}
\end{figure*}

\IEEEPARstart{N}{euroimaging} and neurophysiological modalities provide complementary views of the brain by capturing structural, functional, and electrophysiological information across different spatial and temporal scales. These modalities include structural magnetic resonance imaging (MRI), diffusion MRI techniques such as diffusion tensor imaging (DTI), and functional MRI (fMRI). They also include non-invasive electrophysiological recordings such as electroencephalography (EEG) and magnetoencephalography (MEG), as well as invasive electrophysiological techniques such as electrocorticography (ECoG)~\cite{bandettini2009s}. These complementary measurements enable comprehensive characterization of brain organization, neural dynamics, and disease-related alterations, providing quantitative measures and candidate biomarkers for neurological research and clinical studies~\cite{yen2023exploring}.

Across these distinct modalities, measurements can be modeled as structured matrix- or tensor-valued representations, including covariance and correlation matrices, diffusion tensors, and deformation tensors. Unlike conventional data types, these structured representations are governed by inherent mathematical constraints, such as symmetry and positive definiteness, which must be strictly preserved during statistical analysis and learning. Covariance and correlation matrices capture second-order statistical dependencies and are widely used to characterize functional connectivity, whereas diffusion and deformation tensors characterize structural properties such as tissue microstructure and anatomical geometry. Many of these representations naturally satisfy, or can be regularized to satisfy, the symmetry and positive definiteness constraints required for SPD matrices. Examples include EEG spatial covariance matrices for motor imagery decoding and emotion recognition~\cite{muller1999designing,bos2006eeg}, covariance or correlation matrices derived from fMRI BOLD time series for functional connectivity analysis and clinical prediction~\cite{varoquaux2010detection,dadi2019benchmarking}, voxelwise diffusion tensors from DTI~\cite{basser1994mr}, and deformation tensors from MRI registration~\cite{mcrobbie2017mri}. We collectively refer to these data structures as SPD-valued representations of neuroimaging data.

When equipped with a chosen Riemannian metric, the space of SPD matrices becomes a Riemannian manifold (often referred to as the SPD manifold). Operations performed on this manifold inherently preserve the symmetry and positive definiteness of the data. In neuroimaging, such geometric methods emerged prominently through DTI and computational anatomy. Fletcher and Joshi~\cite{fletcher2004principal,fletcher2007riemannian} pioneered the use of affine-invariant Riemannian metrics for the statistical analysis of diffusion tensors. At approximately the same time, Pennec, Fillard, and Ayache~\cite{pennec2006riemannian,pennec2019riemannian} established a comprehensive Riemannian framework for processing manifold-valued tensor data in computational anatomy. These foundational works paved the way for subsequent statistical and learning methods on SPD manifolds.

Over the past decade, the use of SPD-valued frameworks has expanded substantially across other neuroimaging modalities and neurophysiological recordings. In fMRI, covariance or correlation matrices derived from BOLD time series are used to characterize functional connectivity~\cite{varoquaux2010detection}. Mapping these matrices to the tangent space at a chosen SPD reference point enables conventional statistical modeling and supports individual-level prediction and diagnostic classification~\cite{dadi2019benchmarking,pervaiz2020optimising}. In EEG, spatial covariance matrices are analyzed within an affine-invariant geometry, enabling robust Riemannian classifiers for motor imagery decoding~\cite{barachant2011multiclass,barachant2013classification,congedo2017riemannian,Yger2017}. Related geometric approaches have also been applied to invasive ECoG decoding~\cite{yao2022fast}. These widespread developments motivate a systematic review of these SPD-valued representations, their methodological foundations, and their applications across neuroimaging, brain--computer interfaces (BCIs), and beyond.

In this review, we survey two decades of analysis and learning methodologies developed for SPD-valued neuroimaging data, a paradigm we collectively term \SPDL. \SPDL provides a shared methodological perspective in which covariance or correlation matrices, diffusion tensors, and deformation tensors arising from diverse biomedical applications are treated as a common class of SPD-valued representations. These representations share intrinsic geometric structures and can be rigorously analyzed through principles rooted in geometric statistics. Structurally, \SPDL spans two complementary families: \emph{geometric shallow learning} (GSL) and \emph{geometric deep learning} (GDL), which extend shallow and deep learning paradigms to SPD-valued data, respectively. GSL encompasses manifold-valued statistical methods and optimization, such as Riemannian means, tangent-space embeddings, manifold-valued kernels, geodesic regression, and Riemannian optimization algorithms. Conversely, GDL adapts deep architectures by incorporating geometry-aware layers and manifold-preserving operations. Introducing \SPDL as a unified framework bridges decades of geometric statistical modeling with modern neural network architectures, establishing SPD modeling as a methodological bridge between classical neuroimaging analysis and emerging learning paradigms. As summarized in Table~\ref{Tab:literature}, \SPDL encompasses established BCI and neuroimaging tasks (Panels A--B) as well as emerging problem settings enabled by recent advances in AI (Panel C). Fig.~\ref{fig:spd_learning} illustrates the conceptual progression from modality-specific measurements to geometric representations and learning methods. This perspective underscores three defining characteristics of \SPDL:

\begin{table*}[!t]
\caption{
Applications of SPD Matrix Learning in BCIs and Neuroimaging. (A) classical motor and cognitive decoding; (B) cross-sectional and longitudinal population modeling; and (C) emerging AI-driven SPD-valued neuroimaging tasks.
\label{Tab:literature}
}
\centering
\footnotesize
\renewcommand{\arraystretch}{1.10}
\setlength{\tabcolsep}{4.5pt}
\begin{tabular}{@{}l l l l@{}}
\toprule
\rowcolor{black!7}
\textbf{Neuroimaging Task} & \textbf{Modality} & \textbf{Method} & \textbf{References} \\
\midrule
\rowcolor{black!7}
\multicolumn{4}{c}{\textbf{A. Classical Motor and Cognitive Decoding}} \\
\midrule
A.1\, Motor Decoding & EEG & GSL &
\cite{barachant2010common,barachant2011multiclass,barachant2013classification,yger2016riemannian,congedo2017riemannian,rodrigues2018riemannian,yair2019parallel,rodrigues2020dimensionality,lahav2023procrustes,mellot2024physic,bonet2023sliced,congedo2017closed,rodrigues2017dimensionality,davoudi2017dimensionality,samek2013robust,samek2013divergence} \\
 & EEG & GDL &
\cite{suh2021riemannian,ju2022tensor,pan2022matt,ju2023graph,ju2025deep,ju2020federated,kobler_spd_2022,kobler_controlling_2022,wilson2025deep,wang2025gyroatt}\\
 & ECoG & GSL & \cite{yao2022fast} \\
A.2\, Cognitive Decoding & EEG & GSL & \cite{kalunga2016online,kalunga2018transfer, wang2025dynamic} \\
 & & GDL & \cite{wang2021prototype} \\
\midrule
\rowcolor{black!7}
\multicolumn{4}{c}{\textbf{B. Cross-Sectional and Longitudinal Population Modeling}} \\
\midrule
B.1\, Cross-Sectional Population Modeling (Aging) & M/EEG & GSL &
\cite{sabbagh2019manifold,sabbagh2020predictive,engemann2022reusable,collas2024weakly,mellot2023harmonizing,mellot2024geodesic}\\
 & fMRI & GSL &
\cite{qiu2015manifold,you2022geometric,you2025scalable}\\
B.2\, Longitudinal Population Modeling (Disease Progression) & fMRI & GSL &
\cite{varoquaux2010detection,dadi2019benchmarking,pervaiz2020optimising,ng2014transport,ng2015transport,honnorat2024riemannian} \\
 & Cauchy deformation tensor & GSL &
\cite{kim2017riemannian}\\
 & Clinical/imaging measures & GSL &
\cite{gruffaz2021learning,schiratti2017bayesian}\\
  & MRI & GDL &
\cite{jeong_deep_2024}\\
\midrule
\rowcolor{yellow!12}
\multicolumn{4}{c}{\textbf{C. New AI-Driven SPD-Valued Neuroimaging Tasks}} \\[-0.1em]
\midrule
C.1\, Modeling Brain Functional Dynamics & fMRI & GSL &
\cite{dai2019analyzing}\\
 & fMRI & GDL &
\cite{huang2021detecting,dan2022learning,dan2022uncovering,dan2024exploring,dan2025geodynamics}\\
C.2\, Multimodal Fusion & EEG--fMRI & GDL &
\cite{ju2024deep} \\
 & dMRI--fMRI & GDL &
\cite{dan2024exploring,zhoubrainflow2025} \\
C.3\, Generative Modeling of Brain Connectivity & EEG and fMRI & GDL &
\cite{collas2025riemannian} \\
\bottomrule
\end{tabular}
\end{table*}

\begin{enumerate}

\item[i.] Intrinsic Modeling of SPD-Valued Representations: SPD-valued representations naturally arise from covariance and correlation matrices, diffusion tensors, and other structured matrix-valued representations constructed from neuroimaging and neurophysiological data. \SPDL operates directly within the framework of the SPD manifold, where statistical modeling, optimization, and learning can be formulated while preserving the underlying structural constraints.

\item[ii.] Integration of Classical Geometries and Modern Machine Learning: \SPDL unifies the diverse body of geometric statistical approaches previously developed across DTI, computational anatomy, EEG spatial covariance analysis, and fMRI functional connectivity analysis. By providing a shared mathematical foundation, this framework facilitates the translation of classical geometric statistical methods into modern machine learning paradigms.

\item[iii.] Extension to Emerging AI-Driven Neuroimaging Tasks: Through geometric deep learning, \SPDL extends SPD modeling beyond conventional geometric statistics to incorporate modern deep learning paradigms, such as representation learning, generative modeling, and neural dynamical systems. These developments enable emerging neuroimaging applications, such as dynamic brain connectivity modeling, multimodal fusion, and brain network generation.

\end{enumerate}

This review provides a structured overview of the \SPDL paradigm over the past two decades, focusing on SPD-valued representations in neuroimaging and BCIs. We cover geometric statistics alongside geometric shallow and deep learning methods, providing a coherent perspective on manifold-aware matrix analysis. The remainder of this paper is organized as follows. Section~\ref{sec:COVARIANCE_BASED_NEUROIMAGING_DATA} details the construction of various SPD-valued representations from raw neuroimaging modalities, while Section~\ref{sec:GEOMETRIC_PERSPECTIVES} presents the relevant mathematical and geometric background. We then provide a comprehensive taxonomy of GSL and GDL methodologies in Section~\ref{sec:GEOMETRIC_METHODS}. Turning to practical implementations, Section~\ref{sec:APPLICATIONS} surveys key biomedical and BCI applications, Section~\ref{sec:IMPLEMENTATIONS} summarizes available computational toolkits, and Section~\ref{sec:CHALLENGES} discusses major open challenges and future directions.

\section{SPD-VALUED NEUROIMAGING DATA}
\label{sec:COVARIANCE_BASED_NEUROIMAGING_DATA}

This section categorizes two broad classes of SPD-valued neuroimaging constructs. The first comprises second-order matrix representations derived from neural time series, including covariance, correlation, and cross-power spectral density (CPSD) matrices, whereas the second comprises tensor-based representations arising in structural neuroimaging. These two classes form the basis for the primary mathematical formulations considered in the remainder of this review. We also discuss CPSD matrices as complex Hermitian analogues of real covariance matrices; these require either a dedicated Hermitian positive-definite framework or an explicit real-valued encoding before analysis methods restricted to real-valued SPD matrices can be applied.

\subsection*{Second-Order Matrix Representations}

Let \(x(t) \in \mathbb{R}^{n_C}\) denote the neuroimaging signal at time \(t\), where \(n_C\) is the number of spatial channels (e.g., sensors or brain regions). We assume that standard preprocessing steps, such as band-pass filtering, centering, and scaling, have been applied to \(x(t)\). To make window-level second-order statistics interpretable, approximate stationarity is typically assumed within each short temporal window~\cite{congedo2013new}.

A data segment is formed by stacking \(n_T\) consecutive samples into a matrix \(X \in \mathbb{R}^{n_C \times n_T}\),
\(
X := [x(t), x(t+1), \dotsc, x(t+n_T-1)].
\)
For a zero-mean process, the population spatial covariance matrix is defined as
\(
C := \mathbb{E}[x(t)x(t)^\top],
\)
which captures second-order statistical dependencies across spatial channels. For a segment in which each channel has been centered over time, a conventional sample estimate is \(\widehat{C}:=XX^\top/(n_T-1)\). When the underlying process is nondegenerate, the population covariance is strictly positive definite. In practice, however, empirical covariance estimates may be only positive semidefinite and can be rank-deficient, especially when \(n_T \leq n_C\) or when the signals exhibit strong linear dependence or narrow-band structure. Covariance estimation is therefore often regularized, for example through shrinkage methods such as Ledoit--Wolf or oracle-approximating shrinkage, to improve conditioning and produce stable estimates for subsequent statistical analysis and learning. A more detailed discussion of covariance estimation and regularization is provided in Section~\ref{sec:CHALLENGES}.

The construction and interpretation of these second-order representations depend on the neuroimaging or recording modality. In electrophysiological recordings such as EEG, MEG, and ECoG, spatial covariance is computed directly from multichannel neural time series and summarizes second-order structure in the measured signals. In fMRI, by contrast, covariance or correlation matrices are commonly estimated from preprocessed, spatially aggregated BOLD time series and used to represent functional connectivity~\cite{thirion2014fmri}. As a result, while covariance matrices arise naturally across modalities, their practical role and neuroscientific interpretation may differ depending on the acquisition setting.

When second-order dependence varies with frequency, it is more naturally represented by cross-power spectral density (CPSD) matrices than by a single broadband time-domain covariance matrix. Because CPSD matrices are generally complex Hermitian rather than real symmetric, their geometric treatment requires either a Hermitian positive-definite formulation after full-rank regularization or an explicit real-valued encoding for methods restricted to real SPD matrices. We next review these second-order matrix representations in greater detail.

\begin{itemize}[\setlength{\topsep}{2pt}\setlength{\partopsep}{0pt}\setlength{\itemsep}{0.15\baselineskip}\setlength{\parsep}{0pt}]
    \item EEG/MEG/ECoG Spatial Covariance. EEG and MEG are non-invasive neurophysiological modalities that measure electrical potentials and magnetic fields generated by neural activity, respectively~\cite{teplan2002fundamentals,da2013eeg}. ECoG, by contrast, is an invasive intracranial recording technique in which electrodes are placed directly on the cortical surface~\cite{buzsaki2012origin}. The spatial resolution of EEG and MEG source estimates depends strongly on sensor geometry, head modeling, inverse methods, signal-to-noise ratio, and source configuration. ECoG generally provides higher local spatial resolution because electrodes are placed directly on the cortical surface, albeit at the cost of surgical implantation. Across these modalities, spatial covariance matrices summarize pairwise statistical dependencies between recorded signals, in sensor/electrode space or, for EEG and MEG, in source space after inverse modeling. These covariance representations are widely used for decoding and, with appropriate estimators and confound control, connectivity analysis.
    \item fMRI Functional Connectivity. fMRI is a non-invasive neuroimaging modality that measures brain activity indirectly through the blood-oxygen-level-dependent (BOLD) signal, thereby enabling the study of large-scale functional organization of the brain~\cite{buckner2013opportunities}. Functional connectivity in fMRI is commonly defined as the temporal dependence between BOLD signal fluctuations in spatially distinct brain regions~\cite{friston1994functional}. These dependencies are typically summarized by covariance or correlation matrices, yielding functional-connectome representations.
    \item CPSD of Neural Signals. For a multivariate neural time series, the CPSD matrix at each frequency is complex Hermitian and positive semidefinite; its diagonal entries give channel-wise power, whereas its off-diagonal entries retain cross-spectral magnitude and phase. It therefore describes band-specific cross-spectral structure that broadband covariance does not resolve. Given two time series \(x(t)\) and \(y(t)\), the CPSD at frequency \(f\) is defined as the Fourier transform of their cross-covariance function, \(S_{xy}(f)=\int_{-\infty}^{\infty} R_{xy}(\tau)\,e^{-i2\pi f\tau}\,d\tau\), where \(R_{xy}(\tau)=\mathbb{E}\!\big[x(t+\tau)\,\overline{y(t)}\big]\) denotes the cross-covariance at lag \(\tau\). In practice, CPSD is commonly estimated using Welch's method, which reduces estimator variance by averaging periodograms across overlapping data segments.
\end{itemize}

\subsection*{Tensor-Based Representations}

Tensor-based data structures include diffusion tensors obtained from diffusion tensor imaging (DTI) and deformation tensors arising from MRI registration. These constructs provide local, SPD-valued representations of tissue microstructure and anatomical deformation, respectively.

\begin{itemize}[\setlength{\topsep}{2pt}\setlength{\partopsep}{0pt}\setlength{\itemsep}{0.15\baselineskip}\setlength{\parsep}{0pt}]
    \item DTI Diffusion Tensors. DTI is a diffusion MRI technique that models the local three-dimensional diffusivity of water in tissue and provides microstructural information such as the dominant local orientation of coherent white-matter fibers~\cite{basser1994mr,stejskal1965spin}. At each voxel, DTI estimates a \(3\times 3\) symmetric diffusion tensor \(D\), which captures the anisotropic diffusivity of water and is treated as positive definite when its fitted eigenvalues are strictly positive; regularization may be required when noise produces nonpositive eigenvalues. The tensor is typically estimated from diffusion-weighted MRI signals through the Stejskal--Tanner equation \(S_i = S_0 \exp(-b_i\, g_i^{\top} D g_i)\), where \(S_0\) denotes the signal without diffusion weighting, \(S_i\) the measured diffusion-weighted signal, \(b_i\) the diffusion-weighting factor for the \(i\)th measurement, and \(g_i\) the corresponding diffusion-gradient direction. The eigenvalues and eigenvectors of \(D\) describe the principal diffusivities and their associated orientations, respectively, and thus provide a quantitative characterization of local tissue microstructure.
    \item MRI Deformation Tensors. MRI provides detailed anatomical images~\cite{mcrobbie2017mri}. Structural MRI analyses identify regions in which anatomical measures, such as gray-matter density and cortical thickness, differ across clinical groups or vary with demographic and biological variables. To quantify deformation at a voxel \(u\), registration to a population template yields a deformation field with Jacobian matrix \(J(u)\). The absolute determinant \(\lvert\det J(u)\rvert\) measures local volume change. If the deformation is locally diffeomorphic, \(J(u)\) is nonsingular and the right Cauchy--Green deformation tensor \(C(u):=J(u)^\top J(u)\) is SPD. Its unique SPD square root \(S(u):=C(u)^{1/2}\), the right stretch tensor, encodes local volume change and directional stretching. Following the terminology of the cited neuroimaging literature, we refer to this SPD descriptor as the Cauchy deformation tensor.
\end{itemize}

\section{GEOMETRIC PERSPECTIVES}\label{sec:GEOMETRIC_PERSPECTIVES}

The set of SPD matrices forms a smooth manifold. Equipping this manifold with an appropriate Riemannian metric makes it a Riemannian manifold and enables geometric statistical analysis of SPD-valued neuroimaging data. In what follows, we review the Riemannian foundations of SPD data, introduce representative metrics on SPD manifolds, and summarize the concepts from geometric statistics used in the later application sections. For comprehensive background material on Riemannian geometry, see~\cite{chen1999lectures, petersen2006riemannian, jost2008riemannian}.

\subsection{Riemannian Geometry}

Riemannian geometry is the branch of differential geometry concerned with smooth manifolds equipped with a Riemannian metric, which specifies how to measure distances, angles, and curvature. A smooth manifold \(\mathcal{M}\) is a topological manifold---locally homeomorphic to Euclidean space---equipped with a compatible smooth atlas. At each point \(p\in\mathcal{M}\), the tangent space \(T_p\mathcal{M}\) is a finite-dimensional vector space that provides a linear approximation of the manifold near \(p\). A vector field on \(\mathcal{M}\) is a smooth section \(X:\mathcal{M}\to T\mathcal{M}\) of the tangent bundle, satisfying \(X(p)\in T_p\mathcal{M}\) for every \(p\in\mathcal{M}\).

A Riemannian manifold \((\mathcal{M},g)\) is a smooth manifold endowed with a smoothly varying inner product on each tangent space. Formally, a Riemannian metric is a smooth assignment \(g_p:T_p\mathcal{M}\times T_p\mathcal{M}\to\mathbb{R}\) that is symmetric and positive definite, such that the function \(p\mapsto g_p(X_p,Y_p)\) is smooth for any smooth vector fields \(X\) and \(Y\). The metric allows one to define geometric quantities such as curve lengths, angles, areas, and volumes.

To compare tangent vectors at different points and to define derivatives of vector fields, one introduces a Riemannian connection. The Levi-Civita connection \(\nabla_X Y\) describes how a vector field \(Y\) varies along another vector field \(X\), and it is uniquely determined on a Riemannian manifold by linearity, compatibility with the Riemannian metric, and vanishing torsion. Given a smooth curve \(\gamma:[0,1]\to\mathcal{M}\), a vector field \(X\) along \(\gamma\) is said to be parallel if \(\nabla_{\dot{\gamma}(t)}X = 0\) for all \(t\). For any \(v\in T_{\gamma(0)}\mathcal{M}\), there exists a unique parallel vector field \(X\) along \(\gamma\) satisfying \(X(0)=v\). The terminal vector \(X(1)\) is called the parallel transport of \(v\) along \(\gamma\), and this defines a linear isomorphism
\(
P_\gamma : T_{\gamma(0)}\mathcal{M} \longrightarrow T_{\gamma(1)}\mathcal{M}.
\) A geodesic is a curve \(\gamma:[0,1]\to\mathcal{M}\) whose tangent field is parallel to itself, i.e., \(\nabla_{\dot{\gamma}}\dot{\gamma} = 0\). The Riemannian distance between two points \(p,q\in\mathcal{M}\) is defined as the infimum of the lengths of all piecewise smooth curves joining them:
\(
d(p,q) := \inf_{\gamma} \int_0^1 \|\dot{\gamma}(t)\|_g\, dt,
\)
where the infimum is taken over all curves \(\gamma\) with \(\gamma(0)=p\) and \(\gamma(1)=q\).

The exponential map provides a convenient way to characterize geodesics and their local behavior. For any \(v\in T_p\mathcal{M}\), there exists a unique geodesic \(\gamma_v\) satisfying \(\gamma_v(0)=p\) and \(\dot{\gamma}_v(0)=v\) on a sufficiently small interval. Whenever this geodesic is defined at \(t=1\), the exponential map at \(p\) is defined by \(\exp_p(v) = \gamma_v(1)\), and it satisfies \(\exp_p(0_p)=p\), where \(0_p\) is the zero vector in \(T_p\mathcal{M}\). There exists a neighborhood \(\mathcal{U}\subset T_p\mathcal{M}\) of \(0_p\) on which \(\exp_p\) is a diffeomorphism onto its image \(\mathcal{V}\subset\mathcal{M}\); this image is referred to as an exponential neighborhood of \(p\). More precisely, the injectivity radius at \(p\) is the supremum of all radii \(r>0\) such that \(\exp_p\) restricts to a diffeomorphism from the open ball \(B(0_p,r)\subset T_p\mathcal{M}\) onto its image. In general, the exponential map is locally defined and may not be globally invertible. However, if \(\mathcal{M}\) is geodesically complete (i.e., every geodesic extends indefinitely), then \(\exp_p\) is globally defined for all \(p\in\mathcal{M}\). The inverse of the exponential map, when it exists, is called the logarithmic map: for every \(q\) sufficiently close to \(p\), the logarithmic map is the unique tangent vector \(v\in T_p\mathcal{M}\) satisfying
\(\log_p(q)=v\) whenever \(\exp_p(v)=q\).

\subsection{SPD Manifolds}

The space of $n \times n$ real-valued SPD matrices, denoted by $\mathcal{S}_{++}^n := \left\{ S \in \mathbb{R}^{n \times n} \;\middle|\; S^\top = S, \; x^\top S x > 0, \; \forall x \in \mathbb{R}^n \setminus \{\mathbf{0}\} \right\}$, forms an $n(n+1)/2$-dimensional smooth manifold. This space admits multiple choices of Riemannian metrics, each inducing a distinct geometric structure. Below, we review four representative Riemannian metrics commonly employed in the analysis and learning of SPD-valued representations:

\begin{itemize}

\item Affine-Invariant Riemannian Metric (AIRM)~\cite{pennec2006riemannian}.
AIRM endows the SPD manifold with a geometry that is invariant under affine congruence transformations. Specifically, for any nonsingular matrix \(W \in GL(n)\) and SPD matrices \(P, Q \in \mathcal{S}_{++}^n\), the geodesic distance satisfies
\(
d_{g^{\mathrm{AIRM}}}(WPW^\top,\, WQW^\top)=d_{g^{\mathrm{AIRM}}}(P,Q).
\)
It is also invariant under common positive scaling, i.e., \(d_{g^{\mathrm{AIRM}}}(aP,aQ)=d_{g^{\mathrm{AIRM}}}(P,Q)\) for any \(a>0\).
Formally, AIRM is defined at \(P \in \mathcal{S}_{++}^n\) by
\(
g^{\mathrm{AIRM}}_P(v,w)
:= \left\langle P^{-1/2} v P^{-1/2},\, P^{-1/2} w P^{-1/2}\right\rangle_{\mathcal{F}},
\)
for tangent vectors \(v,w \in T_P\mathcal{S}_{++}^n\).
Geometrically, \((\mathcal{S}_{++}^n, g^{\mathrm{AIRM}})\) is a Hadamard manifold with nonpositive sectional curvature, so any two SPD matrices are connected by a unique geodesic and the Fréchet mean of a finite set of SPD matrices exists and is unique. For dense \(n\times n\) matrices, AIRM-based distance, logarithmic-map, and exponential-map evaluations require \(O(n^3)\) matrix factorizations or spectral operations. In addition, the AIRM Fréchet mean generally has no closed-form expression and is typically computed iteratively, for example using Karcher-flow updates~\cite{karcher1977riemannian,moakher2005differential}.

\item Log-Euclidean Metric (LEM)~\cite{arsigny2005fast,arsigny2006log}. LEM simplifies computation by exploiting the matrix logarithm to map the SPD manifold diffeomorphically onto the Euclidean vector space \(\mathcal{S}^n\) of symmetric matrices. The mapping
\(
\log:\mathcal{S}_{++}^n\longmapsto \mathcal{S}^n
\)
is a global diffeomorphism, and the LEM is defined as the pullback of the Euclidean metric through this matrix-logarithm map. Formally, for any \(P \in \mathcal{S}_{++}^n\) and tangent vectors \(v,w \in T_P\mathcal{S}_{++}^n\),
\(
g^{\mathrm{LEM}}_P(v,w)
= \left\langle D_P\log\, v,\; D_P\log\, w \right\rangle_{\mathcal{F}},
\)
where \(D_P\log\) denotes the differential of the logarithm at \(P\).
LEM induces a flat log-Euclidean geometry in the matrix-log domain. In terms of transformation properties, it is invariant under orthogonal congruence transformations and under common positive scaling, and it is homogeneous under common matrix powers; in particular, one has \(d_{\mathrm{LEM}}(P^p,Q^p)=|p|\,d_{\mathrm{LEM}}(P,Q)\). These properties make LEM computationally convenient: distances, means, and interpolants have closed-form expressions in log coordinates, although evaluating the required matrix logarithms and exponentials still involves cubic-cost spectral operations.

\item Log-Cholesky Metric (LCM)~\cite{lin2019riemannian}. LCM builds upon the Cholesky decomposition \(P=LL^\top\), where \(L\) is a lower-triangular matrix with positive diagonal entries. Let \(\mathcal{L}_{+}\) denote the space of lower-triangular matrices with positive diagonal entries, also called the Cholesky space. The map
\(
\varphi:\mathcal{L}_{+}\rightarrow \mathcal{S}_{++}^n, \varphi(L)=LL^\top,
\)
is a smooth bijection from \(\mathcal{L}_{+}\) onto the SPD manifold \(\mathcal{S}_{++}^n\). For \(P=LL^\top \in \mathcal{S}_{++}^n\), the metric is defined by
\(
g_P^{\mathrm{LCM}}(v,w)
:= \bar{g}_L\!\left(L\big(L^{-1}vL^{-\top}\big)_{\triangle},\,
L\big(L^{-1}wL^{-\top}\big)_{\triangle}\right),
\)
where \((\cdot)_{\triangle}\) extracts the lower-triangular part of a square matrix and scales its diagonal entries by a factor of \(\tfrac12\). The metric \(\bar{g}_L\) on \(\mathcal{L}_{+}\) is given by
\(
\bar{g}_L(X,Y)=\sum_{i>j}X_{ij}Y_{ij}+\sum_{j=1}^{n}X_{jj}Y_{jj}L_{jj}^{-2}.
\)
Unlike AIRM and LEM, LCM depends on the chosen Cholesky parameterization and is not invariant under general congruence transformations. Geometrically, it yields a globally flat Euclidean structure in log-Cholesky coordinates. Because its computations rely on Cholesky factors and elementwise logarithms of their diagonal entries, LCM avoids general eigendecompositions and matrix logarithms. It therefore often offers lower practical computational cost and favorable differentiability properties, making it particularly useful in optimization and deep learning settings.

\item Bures--Wasserstein Metric (BWM)~\cite{bhatia2019bures, malago2018wasserstein}. BWM originates from quantum information theory and optimal transport, and can be viewed as the geometry induced on SPD matrices when they are interpreted as covariance matrices of centered Gaussian distributions. Formally, for \(P \in \mathcal{S}_{++}^n\), it is defined by
\(
g^{\mathrm{BW}}_P(V,W)=\tfrac12\operatorname{tr}\!\left(\mathcal{L}_P[V]\,W\right),
\)
where \(V,W \in T_P\mathcal{S}_{++}^n\) are symmetric tangent matrices, and \(\mathcal{L}_P\) denotes the Lyapunov operator associated with \(P\), i.e., the linear map that assigns to each symmetric matrix \(V\) the unique symmetric solution \(X\) of the Lyapunov equation
\(
PX+XP=V.
\)
This formulation induces the corresponding Bures--Wasserstein geodesic distance in closed form. In terms of transformation properties, BWM is invariant under orthogonal congruence transformations and is homogeneous under common positive scaling, with \(d_{\mathrm{BW}}(aP,aQ)=\sqrt{a}\,d_{\mathrm{BW}}(P,Q)\) for \(a>0\). Geometrically, it endows \(\mathcal{S}_{++}^n\) with a Riemannian structure of nonnegative sectional curvature that is linked to optimal transport and is especially relevant for covariance interpolation, Gaussian modeling, and generative modeling. Its computational cost remains cubic in matrix dimension for dense matrices, typically involving matrix square roots or related spectral operations, but it retains closed-form expressions for both distances and geodesics.

\end{itemize}

\begin{table*}[!t]
\centering
\caption{Comparison of Riemannian metrics with respect to invariance and scaling properties, geometry, closed-form solutions, computational complexity, and representative neuroimaging applications. Definitions of invariance and scaling properties are provided in the main text. Computational complexity estimates are based on \(n \times n\) matrix representations.}
\label{tab:spd_metric_comparison}
\renewcommand{\arraystretch}{1.12}
\setlength{\tabcolsep}{3pt}
\footnotesize

\begin{tabularx}{\textwidth}{
>{\raggedright\arraybackslash}p{0.9cm}
>{\raggedright\arraybackslash}p{4.8cm}
>{\raggedright\arraybackslash}p{3.0cm}
>{\centering\arraybackslash}p{2.0cm}
>{\raggedright\arraybackslash}p{4.0cm}
>{\raggedright\arraybackslash}X}
\toprule
\rowcolor{black!7}
\textbf{Metric} &
\textbf{Invariance/Scaling} &
\textbf{Geometric characteristics} &
\textbf{Closed-form distance and geodesic} &
\textbf{Computational cost} &
\textbf{Uses} \\
\midrule

AIRM &
affine congruence-invariant; scale invariance&
Hadamard; nonpositive sectional curvature &
Yes &
\(O(n^3)\) spectral/factorization steps; iterative mean &
DTI;\allowbreak{} M/EEG;\allowbreak{} ECoG;\allowbreak{} fMRI \\
\addlinespace[1.5pt]

LEM &
orthogonal congruence-invariant; scale invariance; power homogeneity &
flat in the matrix-log domain &
Yes &
\(O(n^3)\) spectral steps; closed-form mean &
EEG;\allowbreak{} fMRI \\
\addlinespace[1.5pt]

LCM &
parameterization-dependent (Cholesky factorization) &
flat in log-Cholesky coordinates &
Yes & \(O(n^3)\) Cholesky steps; closed-form mean & fMRI;\allowbreak{} MRI \\
\addlinespace[1.5pt]

BWM &
orthogonal congruence-invariant; scale homogeneity & nonnegative sectional curvature; linked to optimal transport &
Yes & \(O(n^3)\) square-root/spectral steps; iterative mean & fMRI \\
\bottomrule
\end{tabularx}
\end{table*}

The choice of an appropriate Riemannian metric strongly shapes the geometric and computational characteristics of the resulting space. These representative metrics reflect distinct trade-offs between geometric structure, invariance properties, and computational efficiency on $\mathcal{S}_{++}^n$, as summarized in Table~\ref{tab:spd_metric_comparison}. Specifically, AIRM provides a canonical affine-invariant Riemannian geometry that endows $\mathcal{S}_{++}^n$ with a Hadamard structure and unique geodesics. However, its basic dense-matrix operations incur a cubic computational cost, and workflows requiring iterative estimation of AIRM Fréchet means introduce additional computational overhead. By contrast, LEM and Log-Cholesky equip $\mathcal{S}_{++}^n$ with flat geometries in transformed spaces and admit closed-form means, thereby simplifying optimization and statistical aggregation. The Bures--Wasserstein metric, in turn, offers a geometry linked to optimal transport and centered Gaussian covariance modeling, making it particularly relevant for covariance interpolation and generative learning tasks. Beyond these standard metrics, alternative geometric structures have been explored, including the Fisher--Rao metric derived from elliptical distributions, power-Euclidean metrics that interpolate between linear and log-domain geometries, and the Thompson metric motivated by the order-theoretic structure of positive cones~\cite{bouchard2024fisher, dryden2009non, sra2012new, mostajeran2024differential}.

\subsection{Geometric Statistics}

Geometric statistics extends classical statistical concepts to manifold-valued data, providing a principled framework for analyzing the non-Euclidean representations that arise in these biomedical applications~\cite{pennec2006intrinsic,zhang2013probabilistic}. In this section, we review three foundational concepts and tools from geometric statistics widely used across BCI and neuroimaging pipelines, namely the Fr\'echet mean, principal geodesic analysis, and geodesic regression.

\begin{itemize}
\item Fr\'echet Mean and Variance~\cite{frechet1948elements,grove1973conjugate}. The Fr\'echet mean extends the Euclidean mean to metric spaces. Under appropriate convexity and uniqueness conditions, the Fr\'echet and Karcher means coincide. In particular, the SPD manifold endowed with the AIRM is a Hadamard manifold, on which the Fr\'echet mean is unique and is therefore commonly referred to as the Karcher mean~\cite{afsari2011riemannian}. Given samples \(\{p_i\}_{i=1}^N \subset \mathcal{M}\), a Fr\'echet mean is any \(\mu \in \operatorname*{arg\,min}_{p\in\mathcal{M}} \frac{1}{N}\sum_{i=1}^N d_g^2(p,p_i)\), and the associated Fr\'echet variance is \(\sigma^2 := \frac{1}{N}\sum_{i=1}^N d_g^2(\mu,p_i)\). These quantities play a central role in manifold-valued statistics. In particular, \((\mu,\sigma^2)\) provide natural analogues of location and dispersion parameters and are widely used in defining Gaussian-like distributions on Riemannian manifolds~\cite{said2017riemannian}.

\item Principal Geodesic Analysis (PGA)~\cite{fletcher2004principal}. Principal geodesic analysis extends PCA to manifold-valued data by identifying low-dimensional geodesic structures that pass through a Fr\'echet mean. Let \(\{p_i\}_{i=1}^N \subset (\mathcal{M},g)\) be manifold-valued samples, and let \(\mu\) denote their Fr\'echet mean. In an intrinsic formulation, the first principal geodesic is selected to maximize the variance explained by projections onto a geodesic through \(\mu\), wherever those projections are uniquely defined; subsequent directions are selected subject to orthogonality in \(T_\mu\mathcal{M}\). In practice, a widely used first-order approximation maps the samples to \(T_\mu\mathcal{M}\). After choosing an orthonormal basis of this tangent space, let \(\xi_i\in\mathbb{R}^d\), where \(d=\dim T_\mu\mathcal{M}\), denote the coordinate vector of \(\log_\mu(p_i)\). The empirical tangent-space covariance is then \(\widehat{\Sigma}_\mu:=\frac{1}{N}\sum_{i=1}^N \xi_i\xi_i^\top\). Its eigenvectors, ordered by decreasing eigenvalue and identified with tangent directions \(\{v_j\}_{j=1}^d\), define subspaces \(V_k:=\mathrm{span}(v_1,\ldots,v_k)\). The corresponding exponential image is \(H_k:=\exp_\mu(V_k)\), wherever the map is defined. On a general manifold, \(H_k\) need not be a totally geodesic submanifold, so this construction is best interpreted locally. When the data are sufficiently concentrated around \(\mu\), tangent-space PCA provides an effective approximation to PGA. Algorithms for more direct computation of principal geodesic structures have also been developed for general and structured Riemannian manifolds~\cite{said2007exact,sommer2010manifold}.

\item Geodesic Regression~\cite{thomas2013geodesic}. Geodesic regression extends linear regression to Riemannian manifolds by modeling the dependence between manifold-valued observations \(y_i \in \mathcal{M}\) and scalar predictors \(x_i \in \mathbb{R}\) along a geodesic curve. Given a base point \(p \in \mathcal{M}\) and a tangent vector \(v \in T_p\mathcal{M}\), the fitted value is \(\widehat{y}_i=\exp_p(x_i v)\). If an explicit noise model is used, the observation can be written as \(y_i=\exp_{\widehat{y}_i}(\epsilon_i)\), where \(\epsilon_i \in T_{\widehat{y}_i}\mathcal{M}\) denotes the model error. The parameters \(p\) and \(v\) are estimated by minimizing the sum of squared Riemannian residuals, \(\arg\min_{p,v}\frac{1}{2}\sum_{i=1}^N d_g\!\left(\exp_p(x_i v),\,y_i\right)^2\). This formulation is the natural analogue of Euclidean linear regression for manifold-valued data; Jacobi fields provide the required exponential-map derivatives.

\end{itemize}

\section{GEOMETRIC METHODS}\label{sec:GEOMETRIC_METHODS}

In this survey, we categorize \SPDL methodologies into two families: GSL and GDL. This section summarizes the core approaches in both paradigms.

\subsection{Geometric Shallow Learning}

We classify GSL methods into two primary paradigms: intrinsic approaches and tangent-space approaches.

\begin{itemize}

\item Intrinsic Approach. Intrinsic approaches operate directly on the geometry of SPD manifolds by defining learning and statistical procedures on the manifold itself. In this setting, geodesic distances quantify similarity, and Karcher means provide manifold-respecting averages. These constructions support clustering, classification, interpolation, and statistical analysis without explicitly flattening the data into Euclidean space. Kernel methods constitute one important class of intrinsic approaches for SPD matrices~\cite{jayasumana2013kernel}. Given a suitable Riemannian distance $d$, a geodesic Gaussian kernel can be defined as \(k(P,Q):=\exp\!\left(-d^2(P,Q)/2\sigma^2\right)\), where $P,Q\in\mathcal{S}_{++}^n$ and $\sigma>0$ is a bandwidth parameter. When such kernels are positive definite, they enable standard kernel methods, including support vector machines, kernel PCA, and kernel $k$-means, while remaining consistent with the underlying manifold geometry. Classical manifold-learning methods, such as Isomap and Laplacian eigenmaps~\cite{tenenbaum2000global,belkin2003laplacian}, have also inspired SPD-specific approaches that learn reduced-dimensional SPD-valued representations from high-dimensional SPD matrices~\cite{harandi2014manifold,harandi2017dimensionality}. In addition to distance-based constructions, information-geometric formulations measure dissimilarity through divergence functions rather than Riemannian metrics~\cite{amari2016information}. For example, decomposable Bregman divergences on the cone of positive-definite matrices induce a dually flat information-geometric structure, thereby supporting divergence-based statistical modeling and inference~\cite{amari2014information}.

\item Tangent-Space Approach. The tangent-space approach provides a practical way to handle SPD manifolds by locally linearizing the geometry around a reference point, typically the Fr\'echet mean~\cite{varoquaux2010detection,barachant2011multiclass}. This construction maps manifold-valued data to a Euclidean vector space, thereby enabling the use of standard machine-learning algorithms without geometry-specific modifications. In contrast to intrinsic methods, which operate directly on Riemannian distances and barycenters, tangent-space methods exploit a local first-order approximation of the manifold while benefiting from highly optimized Euclidean models. Formally, given $Q \in \mathcal{S}_{++}^n$, its mapping to the tangent space at $P$ under the AIRM is given by the logarithmic map \(\log_P(Q)=P^{1/2}\,\log\!\left(P^{-1/2} Q P^{-1/2}\right)\,P^{1/2}\). A commonly used Euclidean representation is then obtained by vectorizing the upper-triangular part: \(z=\mathrm{uvec}\!\left(\log\!\left(P^{-1/2} Q P^{-1/2}\right)\right)\in\mathbb{R}^{n(n+1)/2}\), where $\mathrm{uvec}$ rescales off-diagonal entries by $\sqrt{2}$ so that \(\|z\|_2=\|\log_P(Q)\|_P\). That is, the Euclidean norm of the encoded vector coincides with the affine-invariant Riemannian norm of the corresponding tangent vector.

\end{itemize}

\subsection{Geometric Deep Learning}

We organize the GDL category into two representative methodological families: SPDNet-based architectures and gyrovector formulations.

\begin{itemize}

\item SPDNet-Based Approach. SPDNet is a deep neural network architecture designed for classification with SPD matrix inputs~\cite{huang_riemannian_2017}. Its semi-orthogonal BiMap weights are optimized on Stiefel manifolds using stochastic first-order Riemannian methods~\cite{bonnabel2013stochastic}. Gradients through its matrix-valued operations are computed using matrix-backpropagation techniques~\cite{ionescu2015matrix}. The architecture consists of three fundamental layer types:
\begin{enumerate}
\item[i.] BiMap layer: Given an SPD matrix $S$, this layer performs a bilinear congruence transform $W S W^\top$, where $W$ is a learnable transformation matrix typically required to have full row rank. When $W$ has full row rank, the output remains symmetric positive definite. In practice, $W$ is often constrained to lie on the Stiefel manifold to preserve geometric structure and numerical stability.

\item[ii.] ReEig layer: It thresholds the eigenvalues at a positive lower bound, thereby maintaining strict positive definiteness and improving the numerical conditioning of subsequent spectral operations. For $S=U\operatorname{diag}(\sigma_1,\ldots,\sigma_n)U^\top$, where $U$ is orthogonal, the output is $U\operatorname{diag}\!\left(\max\{\epsilon,\sigma_1\},\ldots,\max\{\epsilon,\sigma_n\}\right)U^\top$, where $\epsilon>0$ is a rectification threshold.

\item[iii.] LogEig layer: It applies the matrix logarithm, yielding a symmetric matrix that can be identified with the tangent space at the identity. For $S = U \Sigma U^\top$, one computes $U\, \log(\Sigma)\, U^\top$, where $\log(\Sigma)$ applies an element-wise logarithm to the diagonal entries of $\Sigma$. The resulting symmetric matrix can be processed by standard Euclidean layers such as linear classifiers or fully connected networks.
\end{enumerate}

The Riemannian batch normalization layer, commonly referred to as RieBN, is a prominent normalization extension developed for SPDNet architectures operating on SPD-valued features~\cite{brooks2019riemannian}. Motivated by batch normalization in Euclidean neural networks, Brooks \emph{et al.} designed this layer to estimate the minibatch Fr\'echet mean during training using an iterative Karcher flow procedure. Structurally, RieBN performs Riemannian centering and learnable re-biasing through congruence transformations, mirroring classical whitening and recoloring operations. Specifically, given a batch of SPD matrices $\{S_i\}_{i=1}^N$ with a batch Fr\'echet mean $M$, each sample is first centered via $\bar{S}_i = \Gamma_{M \mapsto I}(S_i) = M^{-1/2} S_i M^{-1/2}$. A learnable multiplicative bias $G$ is then applied via $\Gamma_{I \mapsto G}(\bar{S}_i) = G^{1/2} \bar{S}_i G^{1/2}$. Here, $I$ denotes the identity matrix, and $G \in \mathcal{S}_{++}^n$ parameterizes the learned SPD reference structure. Kobler \emph{et al.} subsequently developed an adaptive momentum formulation that utilizes one-step Karcher flow approximations of minibatch means to update the running estimates of a dataset-level Fr\'echet mean~\cite{kobler_spd_2022}. Related work further extended this framework to control Fr\'echet variance in addition to re-centering the mean~\cite{kobler_controlling_2022}. These structural normalization methods demonstrate task-dependent improvements in running-statistic estimation, training behavior, and model generalization over conventional baselines.

\item Gyrovector Approach. Beyond SPDNet-based architectures, which manipulate SPD features through congruence and spectral layers before mapping the final representation to Euclidean coordinates, gyrovector formulations offer an alternative. Originally developed for hyperbolic geometry~\cite{ungar2008analytic,ungar2022gyrovector}, the gyrovector framework provides a principled algebraic basis for extending Euclidean-style operations, most notably addition and scalar multiplication, to non-Euclidean settings, thereby supporting geometric deep learning on SPD manifolds~\cite{lopez2021vector}.

A \textit{gyrovector space} is an axiomatically defined algebraic structure built on a \textit{gyrogroup}. In this structure, a \emph{gyration} operator accounts for the non-associativity of the binary operation and enables geometry-aware calculus on curved spaces. A gyrogroup on a set \(G\) is equipped with a binary operation \(\oplus\), an identity element, inverses, and gyration automorphisms satisfying the standard gyrogroup axioms. Its characteristic \textit{gyroassociative law} is
\(
a \oplus (b \oplus c) = (a \oplus b) \oplus \operatorname{gyr}(a, b)c,
\)
where \(\operatorname{gyr}:G\times G\rightarrow\operatorname{Aut}(G)\) and \(\operatorname{Aut}(G)\) denotes the group of automorphisms of \(G\). A gyrovector space augments a gyrocommutative gyrogroup with scalar multiplication satisfying the standard gyrovector-space axioms; the gyration operators compensate for the non-associativity of gyroaddition~\cite{ungar2008analytic,ungar2022gyrovector}.

For concrete matrix manifolds, including SPD and Grassmann manifolds, Nguyen constructed explicit operations such as $\oplus$ and $\otimes$ from the underlying Riemannian geometry. These constructions include operations for the affine-invariant, log-Euclidean, and log-Cholesky geometries on SPD manifolds, as well as for a specific Grassmann-manifold geometry, and the resulting operations satisfy the relevant gyro-axioms. These operations induce a corresponding gyrostructure on each manifold that can be used as a geometry-aware calculus for designing learning models~\cite{nguyen2022gyro,nguyen2022gyrovector}.

Nguyen \emph{et al.} constructed basic operations and gyroautomorphisms on Grassmann manifolds from the orthonormal-basis perspective~\cite{nguyen2023building}. They also generalized key notions from gyrovector spaces, including inner products and gyroangles, to SPD and Grassmann manifolds and studied isometry-based models, or gyroisometries, on these spaces.

Building on these developments, they derived fully connected and convolutional layers for SPD neural networks and developed multinomial logistic regression on appropriate positive-semidefinite manifold models. They also proposed an exact backpropagation method using the Grassmann logarithmic map in a projector-based formulation, enabling graph convolutional networks on Grassmann manifolds~\cite{nguyen2024matrix}. Moreover, GyroAtt extends attention mechanisms to SPD, positive-semidefinite, and Grassmann manifold models by using geodesic distances for scoring and weighted Fr\'echet means for aggregation, yielding an attention formulation applicable across these geometries~\cite{wang2025gyroatt}.

\end{itemize}

\section{BCI AND NEUROIMAGING APPLICATIONS}\label{sec:APPLICATIONS}

This section reviews key applications of \SPDL across a wide range of BCI and neuroimaging tasks over the past two decades. Organized by task category, this section summarizes representative geometric formulations and learning methods in each domain. Table~\ref{Tab:literature} groups the cited studies into three thematic areas: classical motor and cognitive decoding; cross-sectional and longitudinal population modeling; and emerging AI-driven SPD-valued neuroimaging tasks.

\subsection*{A. Classical Motor and Cognitive Decoding}
\addcontentsline{toc}{subsection}{(A) Classical Motor and Cognitive Decoding}

\noindent \emph{A.1 Motor Decoding.} In EEG-based motor imagery classification, EEG is recorded while subjects imagine moving the left or right hand, the feet, or the tongue without overt execution~\cite{neuper_imagery_2005}. MI modulates sensorimotor rhythms, primarily in the $\mu$ and $\beta$ bands, and elicits event-related desynchronization/synchronization (ERD/ERS) over sensorimotor areas, thereby providing discriminative control signals for BCIs~\cite{pfurtscheller1999event,pfurtscheller2001motor}. Because ERD/ERS is expressed as band-limited power modulation, spatial filtering is widely used to improve class separability. Spatial filtering projects multichannel trials $X\in\mathbb{R}^{n_C\times n_T}$ onto a lower-dimensional representation $S=W^\top X$, where $W\in\mathbb{R}^{n_C\times d}$. The learned filters define discriminative linear combinations of channels and can improve signal-to-noise ratio, but they do not by themselves identify or remove volume-conduction effects~\cite{blankertz2011single}.

A classical approach is common spatial patterns (CSP)~\cite{muller1999designing}, which estimates spatial filters from class-wise covariance matrices by maximizing the variance ratio between two conditions, typically through the generalized eigenvalue problem $\Sigma^{+}w_i=\lambda_i\Sigma^{-}w_i$, where $\Sigma^{+}$ and $\Sigma^{-}$ denote normalized class-mean covariances. Features are then extracted from the (log-)variance of the projected signals and used as inputs to a classifier; numerous CSP variants have been successfully applied in online BCI systems~\cite{blankertz2007optimizing}. CSP also admits an information-geometric interpretation: the subspace spanned by the leading filters maximizes the symmetric Kullback-Leibler divergence between Gaussian models of the projected signals, while divCSP extends this formulation through regularization and $\beta$-divergence to derive robust variants within a unified divergence-based framework~\cite{samek2013robust,samek2013divergence}.

Riemannian methods model EEG spatial covariance matrices as points on the SPD manifold and perform classification based on distances to class-wise Fr\'echet means, as in the minimum-distance-to-mean (MDM) classifier under the AIRM geometry~\cite{barachant2011multiclass,barachant2013classification}. This line of work has shown competitive performance across multiple EEG paradigms and has also motivated dimensionality reduction methods for compact SPD-valued representations that preserve intrinsic geometry and discriminative structure, typically through tangent-space or graph-based embeddings~\cite{congedo2017riemannian,congedo2017closed,rodrigues2017dimensionality,davoudi2017dimensionality}.

Geometric deep-learning pipelines often build on SPDNet-style architectures. SPDNet itself operates on SPD inputs through a Stiefel-constrained BiMap layer together with ReEig and LogEig operations~\cite{huang_riemannian_2017}. Subsequent variants extend this design to structured time-frequency representations, graph relations among irregular segments, metric learning, attention mechanisms, and learnable filter banks~\cite{ju2022tensor,ju2023graph,suh2021riemannian,pan2022matt,wilson2025deep,wang2025gyroatt}. End-to-end pipelines additionally place learnable convolutional front ends before covariance pooling and SPD processing, as illustrated by TSMNet~\cite{kobler_spd_2022}, manifold-attention models with convolutional preprocessing~\cite{pan2022matt}, and lightweight hybrids combining Gabor wavelet features with SPDNet~\cite{paillard2025green}.

EEG nonstationarity further motivates domain adaptation on SPD manifolds~\cite{lotte2018review}. Representative strategies include recentering covariance matrices around a common reference mean~\cite{zanini2017transfer}, parallel transport across tangent spaces~\cite{yair2019parallel}, Riemannian Procrustes alignment and its dimensionality-transcending variants~\cite{rodrigues2018riemannian,lahav2023procrustes,bleuze_tangent_2022,rodrigues2020dimensionality}, physics-informed alignment for heterogeneous channel layouts~\cite{mellot2024physic}, optimal-transport-based discrepancies and deep log-Euclidean alignment~\cite{bonet2023sliced,ju2025deep}, and domain-specific Riemannian batch normalization for manifold-aware centering and dispersion scaling~\cite{kobler_spd_2022}.

These geometric decoding formulations also extend beyond EEG\@. In ECoG, spatial covariance matrices can likewise be modeled as points on an SPD manifold, and sparse tangent-space feature selection has been shown to support compact geometric representations and accurate decoding of multi-finger movements~\cite{waldert2009review,yao2022fast}.

\noindent \emph{A.2 Cognitive Decoding.} EEG is widely used for cognitive and affective decoding, supporting applications such as emotion recognition, steady-state visually evoked potential (SSVEP) decoding, and dynamic modeling of functional brain activity. In affective neuroscience, frontal alpha asymmetry has been extensively studied in relation to emotional valence and individual differences~\cite{jenke2014feature,bos2006eeg}. Lower alpha power over the left frontal cortex than over the right, often interpreted as greater relative left-frontal cortical activation, has been associated with positive affect, although the direction and robustness of this relationship depend on individual and experimental factors~\cite{bos2006eeg}. In EEG-based emotion recognition, manifold-based methods have been developed to learn discriminative SPD-valued representations in the presence of recording variability. For example, an SPDNet-based framework with domain adaptation and prototype learning has been shown to improve cross-condition emotion decoding on datasets such as DREAMER and DEAP~\cite{wang2021prototype}.

Riemannian methods have also been applied to SSVEP decoding. In SSVEP paradigms, visual stimuli flickering at distinct frequencies elicit phase-locked oscillatory responses with spectral components tied to the attended stimulus frequency and its harmonics~\cite{vialatte2010steady}. The decoding task is therefore to identify the attended stimulus from multichannel EEG. Canonical correlation analysis is a standard approach because it matches neural responses to sinusoidal reference templates~\cite{lin2006frequency}. Recent Riemannian formulations improve robustness to cross-session and cross-subject variability in synchronous, asynchronous, and online settings, and can be further combined with transfer learning to enhance generalization on benchmark datasets~\cite{kalunga2016online,kalunga2018transfer}.

Beyond static decoding, Riemannian modeling has also been used to characterize the dynamics of EEG covariance time series. A recent Riemannian state-space model represents covariance evolution through a latent low-dimensional SPD state and models stochastic variation with a Riemannian Gaussian distribution~\cite{wang2025dynamic}. Inference uses a Riemannian particle filter, whereas parameter estimation uses a Riemannian EM algorithm. The method was evaluated on simulated data and EEG-based emotion datasets~\cite{wang2025dynamic}.

\subsection*{B. Cross-Sectional and Longitudinal Population Modeling}
\addcontentsline{toc}{subsection}{(B) Cross-Sectional and Longitudinal Population Modeling}

\noindent \emph{B.1 Cross-Sectional Population Modeling (Aging).} A widely used paradigm for cross-sectional population modeling is brain-age prediction, in which a model is trained on single-session neuroimaging data to estimate chronological age. The difference between brain-predicted age and chronological age, commonly termed the \emph{brain-age delta}, summarizes deviation from a fitted normative aging pattern but is not a direct measure of biological age; its interpretation depends on calibration, cohort composition, and control of confounding variables~\cite{dadi2021population}.

Across the lifespan, the brain undergoes structural and functional reorganization, including volumetric decline. Cognitive performance also changes with age. Resting-state neuroimaging studies often report reduced within-network coherence, whereas between-network coupling may reflect altered functional segregation; however, the magnitude and even the direction of these effects depend on the region and system under study rather than following a universally monotonic pattern~\cite{ferreira2013resting,cole2017predicting,puxeddu2020modular}. Because aging affects both fast electrophysiological dynamics and slower hemodynamic responses, M/EEG and fMRI provide complementary views for characterizing lifespan-related network changes~\cite{engemann2020combining}.

For M/EEG-based modeling, regularized spatial covariance matrices provide geometry-aware representations that can be relatively robust to anatomical variability without requiring explicit source localization~\cite{sabbagh2019manifold,sabbagh2020predictive}. To reduce domain shifts across sessions, subjects, and acquisition sites, geometric alignment procedures, including tangent-space recentering, scaling, and controlled rotations, have been developed on the Riemannian manifold of covariance matrices. Recent harmonization frameworks jointly estimate geometric transformations and predictive mappings and have achieved competitive predictive performance across multiple datasets, with improved robustness under heterogeneous recording conditions~\cite{mellot2023harmonizing,collas2024weakly,mellot2024geodesic}.

In fMRI-based aging studies, sparse inverse covariance estimation has been used to estimate SPD precision matrices that encode conditional functional connectivity. Under the log-Euclidean metric, applying the matrix logarithm yields Euclidean log-domain representations while retaining the metric's geometric structure and computational tractability. Statistical learning models such as ridge regression can then be applied to these log-domain features to characterize lifespan-related functional reorganization, including cortical-subcortical integration and altered relationships within and between large-scale systems~\cite{qiu2015manifold}.

Geometric modeling has also been extended from aging-specific functional connectivity analysis to the analysis of correlation matrices derived from a broader range of neuroimaging modalities. Although correlation matrices are widely used to characterize interregional statistical relationships, Euclidean treatments often ignore their global geometric structure. Full-rank correlation matrices form a constrained subset of the SPD manifold characterized by a unit diagonal. Standard SPD-manifold operations do not generally preserve this unit-diagonal constraint and can introduce numerical and computational difficulties in high dimensions. Recent work addresses these issues through diffeomorphic embeddings that map correlation matrices to Euclidean coordinates while implicitly preserving manifold constraints, thereby enabling scalable analyses within standard machine-learning pipelines. Across the reported applications, including resting-state fMRI fingerprinting, behavioral prediction, and EEG hypothesis testing, these embeddings improved computational efficiency; in the predictive tasks, they also achieved competitive or improved performance relative to the evaluated baselines~\cite{you2022geometric,you2025scalable}.

\noindent \emph{B.2 Longitudinal Population Modeling (Disease Progression).} Functional connectivity derived from fMRI has been widely used to study neuropsychiatric disorders, including autism spectrum disorder, depression, and schizophrenia~\cite{yahata2016small,ichikawa2020primary,yoshihara2020overlapping}, as well as neurodegenerative conditions such as Alzheimer's and Parkinson's diseases~\cite{filippi2019resting}.

A common geometric approach models covariance or correlation matrices used to represent functional connectivity as points on the SPD manifold and applies a tangent-space embedding at the Fr\'echet mean under the AIRM~\cite{varoquaux2010detection}. This local linearization yields Euclidean features that can be used with standard machine-learning models, and empirical studies on multiple fMRI datasets have shown competitive predictive performance for tangent-space representations~\cite{dadi2019benchmarking}. Related benchmark analyses have further examined the impact of parcellation, connectivity estimation, confounds, classifiers, and prediction strategies on functional-connectivity-based prediction using large resting-state datasets such as the Human Connectome Project and UK Biobank~\cite{pervaiz2020optimising}.

Riemannian formulations have also been used to harmonize functional connectivity while preserving the SPD structure of the covariance or correlation matrices that represent it. A typical strategy maps these matrices to the tangent space at the population geometric mean, performs statistical correction in the linearized domain, and maps the harmonized data back to the manifold through the exponential map. The cited study reported reduced site-related effects while preserving the biological associations and predictive information evaluated in its experiments~\cite{honnorat2024riemannian}.

Longitudinal neuroimaging models are used to characterize temporal changes associated with disease progression, aging, treatment response, and neurodevelopment~\cite{diggle2002analysis}. In a longitudinal fMRI setting, matrix whitening transport has been introduced to align covariance matrices used to represent functional connectivity on SPD manifolds under Wasserstein geometry~\cite{ng2014transport,ng2015transport}. The cited studies reported that transporting covariance estimates into a common coordinate system reduced statistical dependencies among connectivity features and improved classification stability across longitudinal conditions, such as pre- and post-intervention comparisons.

A more explicit trajectory-based formulation is provided by Riemannian mixed-effects modeling with stochastic inference, which enables nonlinear group trajectory estimation for manifold-valued longitudinal data while accounting for inter-subject spatiotemporal variability~\cite{schiratti2017bayesian}. In this framework, individual trajectories are modeled as exp-parallelizations of a population-average curve on the manifold, yielding a decomposition of variability into spatial effects, which determine trajectory location and direction, and temporal effects, which capture subject-specific time reparameterization. Combined with Bayesian mixed-effects estimation and Monte Carlo inference, this formulation supports joint estimation of population-level disease trajectories and subject-specific random effects from longitudinal neuropsychological scores and manifold-valued observations~\cite{schiratti2017bayesian}.

Related nonlinear mixed-effects models have been extended to longitudinal deformation analysis by encoding voxelwise anatomical changes as Cauchy deformation tensors and modeling subject-specific trajectories on the SPD manifold~\cite{kim2017riemannian}. More flexible geometric disease-course mapping has subsequently been developed through learned Riemannian metrics based on diffeomorphic pushforward constructions in reproducing kernel Hilbert spaces~\cite{gruffaz2021learning}. Recent geometric deep-learning approaches further combine manifold modeling with continuous-time dynamics for Alzheimer's disease progression, integrating trajectory estimation, latent geometric evolution, and monotonicity constraints. In experiments on the TADPOLE dataset, the cited study reported improved longitudinal predictions and produced trajectories that better satisfied its evaluated monotonicity criteria than the selected baselines~\cite{jeong_deep_2024}.

\subsection*{C. New AI-Driven SPD-Valued Neuroimaging Tasks}
\addcontentsline{toc}{subsection}{(C) New AI-Driven SPD-Valued Neuroimaging Tasks}

\noindent \emph{C.1 Modeling Brain Functional Dynamics.} Dynamic functional connectivity (dFC) is commonly represented by a sequence of windowed covariance or correlation matrices computed from BOLD time series, for example using sliding-window correlation. These sequences may then be analyzed by clustering or state analysis in Euclidean space~\cite{allen2014tracking}. Because apparent temporal variation may also arise from finite-sample and windowing effects, temporal variation in such trajectories should be interpreted as putative evidence of dynamic connectivity unless validated against appropriate null models. When these windowed estimates are regularized to remain positive definite, they can be viewed as trajectories on the SPD manifold, which motivates geometric approaches to dynamic connectivity analysis. In this setting, Zhang \emph{et al.} introduced the transported square-root velocity representation (TSRVF) for reparameterization-invariant registration and comparison of covariance trajectories~\cite{zhang2018rate}, while Dai \emph{et al.} further modeled dFC as trajectories on the SPD manifold and developed metric-based dimensionality reduction to obtain compact embeddings that preserve discriminative structure~\cite{dai2019analyzing}. Experiments on Human Connectome Project data indicated that these geometric trajectory representations achieved task-classification performance competitive with or better than the evaluated dFC baselines~\cite{dai2019analyzing}.

Another line of work treats dFC as a candidate marker of brain-state transitions and adopts SPD-valued representations for change detection. SPDNet-based architectures combine BiMap, ReEig, and LogEig layers with recurrent neural networks to learn low-dimensional representations of dFC matrices and model their temporal evolution~\cite{huang2021detecting}. Attention mechanisms have also been incorporated into these manifold-based networks to emphasize state-dependent connectivity patterns while preserving intrinsic geometric structure~\cite{dan2022learning}. Related geometric deep-learning methods have been used to extract shape signatures from resting-state functional-connectivity distributions for brain-state characterization~\cite{dan2022uncovering}. In the reported Human Connectome Project experiments, the change-detection models achieved higher performance than the evaluated conventional learning baselines~\cite{huang2021detecting,dan2022learning}.

State-space models provide an alternative for capturing latent dynamics underlying evolving dFC trajectories. An earlier state-space formulation was developed for event-related fMRI~\cite{faisan2007hidden}, whereas a recent approach combines geometric deep learning with a state-space model whose latent states evolve directly on the SPD manifold~\cite{dan2025geodynamics}. The cited study reported higher state-discrimination performance than Euclidean state-space baselines and associations between the inferred dynamics and cognitive, behavioral, or clinical variables in large-scale fMRI datasets~\cite{dan2025geodynamics}.

\noindent \emph{C.2 Multimodal Fusion.} Multimodal neuroimaging provides complementary information on brain structure and function, motivating fusion strategies that jointly exploit multiple non-invasive modalities~\cite{luo2024multimodal}. A representative example is simultaneous EEG--fMRI, which combines the complementary temporal and spatial sensitivities of the two modalities for functional connectivity analysis~\cite{philiastides_inferring_2021,warbrick_simultaneous_2022}. At the same time, multimodal fusion remains challenging because the modalities differ substantially in sampling rates, spatial sensitivity, and underlying physiological mechanisms~\cite{he2008multimodal}.

Recent geometric approaches address cross-modal integration by representing both modalities in a common manifold-aware latent space. Deep Geodesic Canonical Correlation Analysis (DGCCA), for example, aligns EEG covariance matrices and fMRI covariance or correlation matrices used to represent functional connectivity on the SPD manifold~\cite{ju2024deep}. Rather than assuming matched covariance structure across modalities, DGCCA learns shared latent representations in which cross-modal correspondence is promoted through geodesic correlation on the manifold. The resulting features support multimodal fusion for downstream tasks such as EEG-based motor imagery classification~\cite{ju2024deep}.

Multimodal fusion has also been studied for jointly modeling structural and functional connectivity. Functional connectivity (FC), estimated from BOLD signals, is commonly represented by a covariance or correlation matrix encoding interregional statistical dependence; when full rank or suitably regularized, this matrix can be treated as SPD. Structural connectivity (SC), derived from diffusion MRI tractography, characterizes the anatomical pathways linking distant brain regions~\cite{honey2009predicting}. Geometry-aware fusion methods use the structural connectome to constrain functional representations across multiple scales. In particular, scattering transforms derived from SC can define multiscale harmonic bases for FC representations, and Mixer-style architectures can then aggregate the resulting row-wise and column-wise features to learn multiscale Riemannian embeddings for brain-state characterization and structure--function analysis~\cite{dan2024exploring}.

A complementary strategy models SC--FC coupling as a transport process on an SPD manifold. In the BrainFlow formulation, SC and state-specific FC representations are embedded or regularized within an SPD/Cholesky parameterization before flow matching; this qualification is necessary because a raw tractography adjacency matrix is not automatically SPD. The model maps a static SC representation to multiple state-specific FC representations, while an inverse mapping estimates SC from FC. Consensus constraints promote consistency across inverse mappings that share anatomical substrates~\cite{zhoubrainflow2025}.

\noindent \emph{C.3 Generative Modeling of Brain Connectivity.} Generative modeling of brain connectivity has become increasingly relevant for controlled investigation of population variability, characterization of disease-related network alterations, and data augmentation when labeled neuroimaging datasets are limited~\cite{vanrullen2019reconstructing,akarca2021generative,habashi2023generative}. A central challenge is that covariance and correlation matrices used to represent connectivity occupy constrained matrix spaces and, when positive definite, can be treated as points on the SPD manifold, whereas many generative models are formulated in unconstrained Euclidean spaces and therefore do not explicitly preserve these constraints or their associated geometry~\cite{ju2023score}.

This constraint can be addressed through global diffeomorphisms between constrained matrix manifolds and Euclidean coordinate spaces. Representative transformations include the matrix logarithm for EEG covariance matrices and normalized Cholesky decompositions for fMRI correlation matrices. These transformations induce pullback metrics on the original matrix manifolds, under which Riemannian conditional flow matching is equivalent to standard conditional flow matching after transformation to Euclidean coordinates. Applying the inverse transformations to generated samples enforces the matrix constraints~\cite{collas2025riemannian}. Across the reported EEG and fMRI experiments, the resulting models generated connectivity matrices that more closely matched the empirical distributions and structural properties than the evaluated baselines and supported downstream prediction and disease-related connectivity analysis~\cite{collas2025riemannian}.

\section{COMPUTATIONAL TOOLS}\label{sec:IMPLEMENTATIONS}

Implementing \SPDL pipelines requires tools for constrained optimization, manifold computation, and modality-specific preprocessing. Manifold-valued parameters and orthonormal weight matrices cannot generally be updated by unconstrained Euclidean steps without leaving their feasible sets; they therefore require Riemannian optimization or an appropriate parameterization.

Riemannian optimization provides a natural framework for problems of the form $\min_{x\in\mathcal{M}} f(x)$, where $\mathcal{M}$ is a Riemannian manifold~\cite{absil2008optimization}. A Riemannian gradient descent update takes the form
\(
x_{k+1} = R_{x_k}\!\left(-\alpha_k\,\mathrm{grad}\, f(x_k)\right),
\)
where $\mathrm{grad}\,f(x_k)\in T_{x_k}\mathcal{M}$ denotes the Riemannian gradient and $R:T\mathcal{M}\rightarrow\mathcal{M}$ is a retraction that maps the tangent-space update back to the manifold.

Several open-source toolboxes provide general-purpose support for optimization on manifolds. Frameworks such as \texttt{Manopt} for MATLAB, \texttt{pymanopt} for Python, and \texttt{Manopt.jl} for Julia support finite-dimensional Riemannian optimization and implement a broad range of smooth algorithms, including steepest descent, conjugate gradient, and trust-region methods~\cite{boumal2014manopt,townsend2016pymanpot,bergmann2022manoptjl,axen2023manifoldsjl}. While MATLAB \texttt{Manopt} and Python \texttt{pymanopt} mainly focus on smooth optimization, Julia \texttt{Manopt.jl} also supports certain nonsmooth Riemannian methods, including subgradient and proximal-gradient algorithms.

For deep models with manifold-valued parameters, Riemannian extensions of adaptive optimizers, including Adam, AdaGrad, and AMSGrad, have been developed~\cite{becigneul2018riemannian}. These methods compute Riemannian gradients in tangent spaces and update parameters through retraction or exponential-map steps; methods with momentum additionally transport their state variables between tangent spaces using suitable vector transports. Riemannian stochastic gradient descent was formalized by Bonnabel~\cite{bonnabel2013stochastic}, and later adaptive methods build on this intrinsic formulation. Within the PyTorch ecosystem, \texttt{geoopt}~\cite{kochurov2020geoopt} provides practical implementations of Riemannian optimizers for several manifolds, including the SPD manifold under AIRM\@. By contrast, \texttt{GeoTorch}~\cite{lezcano2019geotorch} does not implement intrinsic Riemannian optimization directly; instead, it enforces manifold constraints through \emph{trivialization}, allowing standard Euclidean optimizers to be applied to unconstrained parameters.

Beyond optimization, several Python libraries support geometric computation on manifolds. \texttt{Geomstats}~\cite{miolane2020geomstats} provides a unified framework for Riemannian geometry and geometric learning, including Fr\'echet means, tangent PCA, geodesic regression, parallel transport, and probabilistic modeling on SPD, Stiefel, and other manifolds. In BCI research, \texttt{pyRiemann}~\cite{barachant2023pyriemann} provides widely used implementations of covariance estimators, tangent-space mappings, and shallow Riemannian classifiers such as the minimum-distance-to-mean (MDM) classifier, together with EEG-specific preprocessing utilities. By comparison, \texttt{MNE-Python}~\cite{gramfort2013meg} and \texttt{Nilearn}~\cite{abraham2014nilearn} primarily focus on preprocessing and conventional statistical or machine-learning workflows for electrophysiological data and fMRI, respectively. Dedicated intrinsic SPD-manifold functionality is generally provided by packages such as \texttt{pyRiemann}, \texttt{Geomstats}, or \texttt{geoopt}.

\texttt{SPD Learn} \texttt{v0.1.0} is an emerging Python package for geometric deep learning on SPD matrices. It provides spectral operators, trivialization-based parameterizations for Stiefel and SPD constraints, and reference implementations of representative SPDNet-based models~\cite{aristimunha2026spd}. Its interfaces are designed to interoperate with BCI, neuroimaging, and machine-learning toolkits for reproducible benchmarking.

\section{CHALLENGES}\label{sec:CHALLENGES}

In this section, we discuss three key technical challenges in \SPDL: covariance estimation, computational complexity, and interpretability. Addressing these issues is important for improving the stability, robustness, and practical applicability of learning frameworks in BCI and neuroimaging. We consider broader issues such as noise and artifacts, protocol heterogeneity, and limited data only insofar as they affect SPD estimation and learning, rather than attempting a comprehensive survey of neuroimaging acquisition and preprocessing.

\noindent \emph{Covariance Estimation.} Reliable covariance estimation remains a central issue in neuroimaging analysis, particularly in \SPDL, where downstream representations and classifiers depend directly on the quality of SPD matrices. In practice, estimation is challenged by high dimensionality, artifacts and outliers, and the temporal nonstationarity of brain signals.

A first difficulty arises from the high-dimensional regime that is common in fMRI functional connectivity~\cite{varoquaux2010brain} and EEG analysis~\cite{engemann2015automated}, where the number of variables may be comparable to or larger than the number of temporal samples. In this setting, empirical covariance matrices are often ill-conditioned or rank-deficient. Regularized estimators are therefore widely used. Shrinkage estimators improve conditioning by combining the sample covariance with a structured target. Examples include Ledoit--Wolf~\cite{ledoit2004well}, which has been shown to improve EEG decoding performance~\cite{lotte2010regularizing}, and oracle-approximating shrinkage (OAS), illustrated in Algorithm~\ref{Alg:fMRI_correlation}~\cite{chen2010shrinkage}. Algorithm~\ref{Alg:fMRI_correlation} follows the standard OAS procedure but omits the closed-form expression for $\lambda$ for concision. In fMRI, residual temporal autocorrelation can reduce the effective sample size and thereby miscalibrate shrinkage~\cite{chen2010shrinkage,lund2006non}. Sparse precision-matrix estimators such as the graphical lasso~\cite{friedman2008sparse} instead encourage a sparse conditional-dependence graph, with zero off-diagonal entries encoding conditional independences under a Gaussian model. The biases of these approaches depend on their tuning and structural assumptions: overly strong sparsity can suppress distributed dependencies or distort hub structure~\cite{varoquaux2012markov}, whereas an overly restrictive shrinkage target can oversmooth heterogeneous covariance structure~\cite{rahim2019population}.

A second challenge arises from artifacts and outliers, such as eye blinks in EEG and head motion in fMRI, as well as from heavy-tailed or otherwise non-Gaussian observations~\cite{Parra2005,power2012spurious}. Such effects can substantially bias empirical covariance estimates and weaken downstream geometric representations. Robust scatter estimators, including Tyler's M-estimator~\cite{tyler1987distribution}, reduce sensitivity to heavy-tailed elliptical observations, although structured physiological and motion artifacts still require appropriate preprocessing or explicit modeling. Classical Tyler estimation typically requires the number of samples to exceed the data dimension. Regularized extensions have been proposed for high-dimensional settings~\cite{sun2014regularized}, but they generally increase computational cost.

A third challenge is temporal nonstationarity. Neuroimaging signals often exhibit time-varying covariance structure, which violates the stationarity assumptions underlying standard estimators. A stationary estimator can then average across distinct covariance regimes, and the resulting model mismatch may propagate to downstream decoding or classification, reducing robustness across sessions, conditions, or subjects~\cite{shenoy2006towards,samek2012stationary,allen2014tracking}.

\begin{algorithm}
\caption{Simplified OAS Correlation Estimation for fMRI Signals}
\label{Alg:fMRI_correlation}
\KwIn{Matrix $\mathbf{X} \in \mathbb{R}^{M \times T}$ of preprocessed regional fMRI time series, where $M$ is the number of regions of interest and $T$ is the number of time points.}
\KwOut{Regularized correlation matrix $\widehat{\mathbf{C}}$.}
\BlankLine
\textbf{Step 1: Normalize regional time series (zero mean and unit $\ell_2$ norm):} Let $\mathbf{x}_m$ denote row $m$ of $\mathbf{X}$. For $m \in \{1,\ldots,M\}$,\\
\hspace*{1em} (1) Subtract the temporal mean: \(\mathbf{x}_m \leftarrow \mathbf{x}_m - \mathrm{mean}(\mathbf{x}_m)\)\;
\hspace*{1em} (2) If the centered row is nonzero, normalize it by the $\ell_2$ norm: \(\mathbf{x}_m \leftarrow \mathbf{x}_m/\sqrt{\sum_{t=1}^{T} x_{m,t}^2}\)\;
\BlankLine
\textbf{Step 2: Compute the sample correlation:} \(\mathbf{C} \leftarrow \mathbf{X}\mathbf{X}^\top\)\;
\BlankLine
\textbf{Step 3: Apply OAS:} Let $\lambda\in[0,1]$ denote the shrinkage coefficient, and compute
\[
\widehat{\mathbf{C}} \leftarrow (1 - \lambda)\mathbf{C}
+ \lambda\,\operatorname{tr}(\mathbf{C})\mathbf{I}_M/M.
\]
\end{algorithm}

\noindent \emph{Computational Complexity.} Computational cost remains a major limitation in \SPDL, with two recurring bottlenecks: Fr\'echet mean estimation and repeated large-scale matrix operations. The burden increases with the number of samples, network depth, and matrix dimensionality.

A first source of overhead is the computation of Fr\'echet means, which frequently appears in manifold-valued statistics and in neural components such as Riemannian normalization layers. For SPD matrices, the cost depends strongly on the underlying metric. Under the LEM, the Fr\'echet mean admits a closed-form expression obtained by averaging matrix logarithms and mapping the result back with the matrix exponential. Under the AIRM, by contrast, no general closed-form solution is available, so the mean is typically estimated iteratively. In the original RieBN, this introduces an inner Karcher-flow loop~\cite{brooks2019riemannian}; later variants use momentum-based or one-step approximations to reduce that cost~\cite{kobler_spd_2022,kobler_controlling_2022}. These approximations reduce runtime but need not recover the exact minimizer of the Fr\'echet functional.

A second bottleneck arises from matrix functions on $\mathcal{S}_{++}^d$, especially in architectures that rely on spectral operators. Among commonly used metrics, AIRM is particularly expensive because geodesic interpolation, parallel transport, and related operations require matrix square roots and inverse square roots. For a batch of $B$ covariance matrices of size $d\times d$, the cost of these operations scales as $O(Bd^3)$. Representative \SPDL architectures incur similar costs: ReEig and LogEig require eigendecompositions for individual matrices, while Riemannian normalization additionally requires mean estimation~\cite{ionescu2015matrix,huang_riemannian_2017,brooks2019riemannian}. For a network with $L$ such operations, the dominant spectral cost scales as $O(BLd^3)$, with backward propagation incurring a comparable asymptotic cost; this estimate ignores constant factors and the number of iterations used for mean estimation.

LEM can reduce repeated manifold-operation costs when a pipeline uses a single log-Euclidean embedding: each SPD matrix requires one spectral decomposition for the matrix logarithm, after which downstream operations in log coordinates are Euclidean. This avoids repeated AIRM square-root and inverse-square-root evaluations, although the initial eigendecomposition and its derivatives still incur an $O(d^3)$ cost.

To reduce this burden, researchers have developed decomposition-free approximations to matrix functions. A representative example is the use of Newton-Schulz iterations to approximate matrix square roots, which replaces eigendecomposition with GPU-friendly matrix multiplications and can improve efficiency in both forward and backward passes~\cite{li_towards_2018}. Such methods, however, remain approximate, and their accuracy depends on the number of iterations and the conditioning of the input matrix.

\begin{figure}[!t]
   \centering
     \includegraphics[width=0.88\columnwidth]{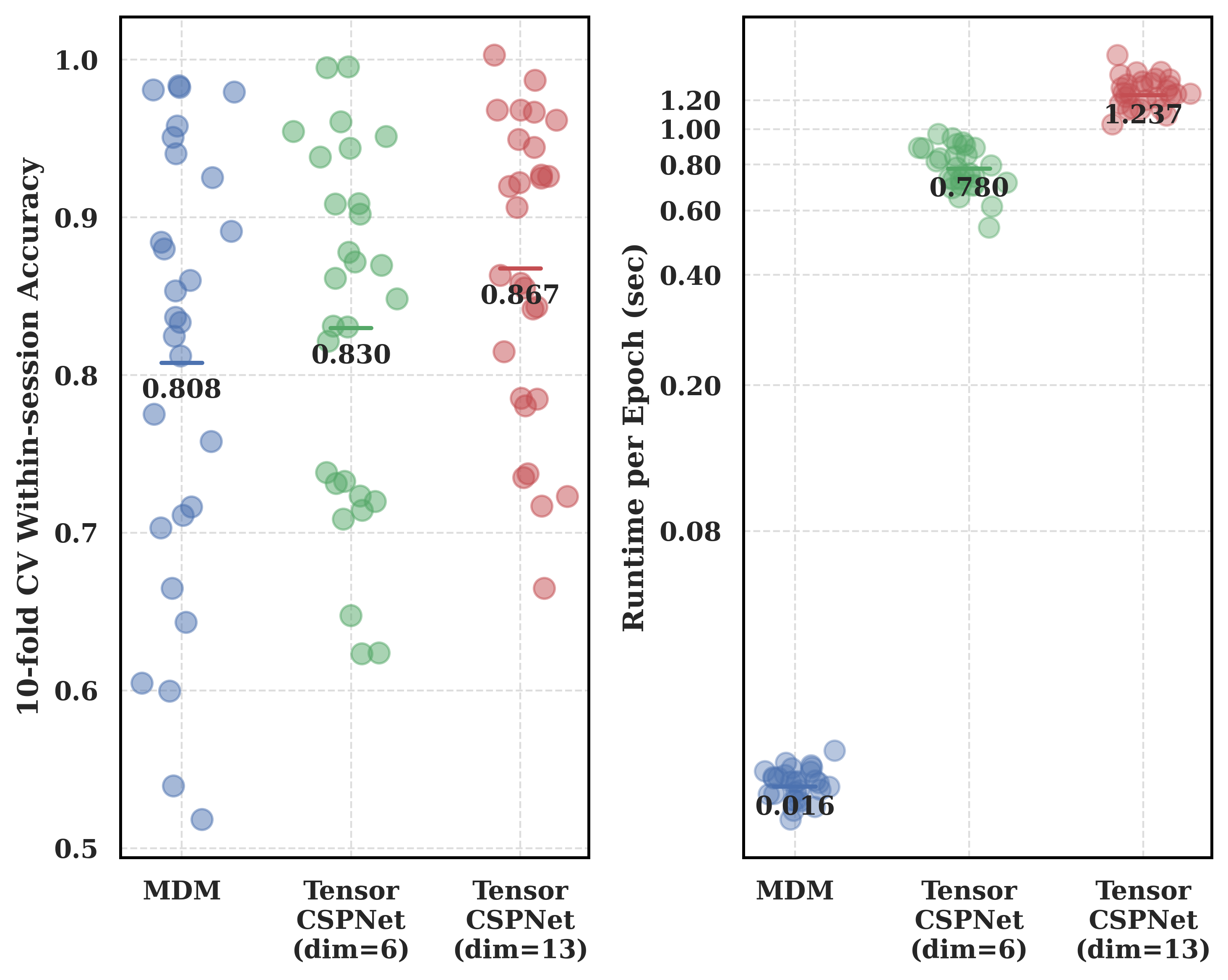}
   \caption{Accuracy--efficiency trade-off on BNCI2015-001~\cite{faller2012autocalibration}. Points denote 10-fold within-session accuracy (left) and per-epoch runtime (right) for MDM and single-BiMap-layer Tensor-CSPNet with dimensions 6 and 13 across 28 sessions (12 subjects); bars and labels show means.}
   \label{fig:dim_accuracy}
   \vspace{-4pt}
 \end{figure}

Another strategy for improving scalability is to reduce the size of covariance matrices through bilinear mappings or manifold-based embeddings before downstream learning~\cite{harandi2014manifold,rodrigues2017dimensionality,harandi2017dimensionality,kalaganis2019riemannian}. While this can substantially lower computational cost, it may also remove discriminative information. For example, in EEG-based motor imagery classification, reducing the SPD dimension lowered per-epoch runtime while producing a modest decrease in mean within-session accuracy in the reported experiment (Fig.~\ref{fig:dim_accuracy})~\cite{ju2022tensor}. In practice, \SPDL therefore involves a recurring trade-off between computational efficiency and representation fidelity.

\noindent \emph{Interpretability.} Interpretability remains a fundamental challenge in \SPDL, particularly in neuroimaging applications where model outputs are expected to support neuroscientific insight rather than prediction alone. Two aspects are especially important: feature-level interpretability, which relates model components to spatial locations, temporal intervals, and frequency bands, and physiological interpretability, which asks whether the extracted features reflect meaningful task-related neural processes rather than incidental artifacts or noise.

A first difficulty concerns spatial mapping. In conventional linear decoding, classifier weights cannot generally be interpreted directly and must be transformed into activation patterns to avoid misleading localization induced by noise suppression~\cite{haufe2014interpretation,Parra2005}. More generally, interpretable neuroimaging models require a principled correspondence between learned parameters and neurophysiologically meaningful representations~\cite{bach_pixel-wise_2015,lundberg_unified_2017,sundararajan_axiomatic_2017,samek_explainable_2019}. In \SPDL, this problem is further complicated by the use of SPD-valued representations and tangent-space parameterizations: model coefficients are often defined on vectorized covariance features rather than directly on sensor-level or voxel-level signals. As a result, they do not admit a straightforward mapping to standard spatial topographies. Recent EEG studies have begun to address this issue by projecting tangent-space weights back to spatial and spectral patterns in modality-specific ways~\cite{xu2020tangent,kobler2021interpretation}.

A second difficulty concerns physiological validity. High predictive performance does not necessarily imply that a model has captured the neural mechanism of interest. In motor-imagery BCIs, for example, discriminative features can be driven by non-task-related physiological artifacts or measurement noise rather than motor-imagery-specific activity~\cite{kobler2021interpretation}. Established physiological signatures, such as contralateral lateralization of sensorimotor activity, can instead provide criteria for assessing whether learned features are neurophysiologically plausible, although their expression varies substantially across subjects~\cite{gonzalezfeature}. Interpretability in \SPDL therefore requires not only identifying where discriminative information is expressed, but also assessing whether the learned representation is biologically plausible.

\section{CONCLUSIONS}

This review has synthesized two decades of geometric statistical and learning methodologies for SPD-valued representations within the unified framework of \SPDL. By modeling covariance and correlation matrices, diffusion tensors, deformation tensors, and other structured representations as points on SPD manifolds when the representations are appropriately constructed or regularized, \SPDL provides a common mathematical framework for data with intrinsic structural constraints. This perspective enables principled geometric statistics and learning while preserving key properties such as symmetry and positive definiteness under valid manifold operations.

We reviewed the geometric foundations of SPD manifolds, organized methodological developments into GSL and GDL, and surveyed applications spanning neural decoding, population modeling, dynamic connectivity, multimodal analysis, and generative modeling. These developments demonstrate that diverse neuroimaging and neurophysiological tasks can be described by a shared statistical formulation despite differences in data sources and scientific objectives. SPD-valued representations therefore serve not only as mathematically structured objects but also as a common representation class connecting established analysis pipelines with emerging AI-driven neuroimaging tasks.

Despite these advances, several challenges remain in the development of \SPDL. Reliable estimation of covariance and connectivity representations under noise, nonstationarity, and limited samples requires further statistical advances. Large-scale matrix operations and manifold optimization also remain computationally demanding for high-dimensional and deep-learning applications. Moreover, improving the interpretability and neuroscientific validation of \SPDL models remains essential. Addressing these challenges will be critical for developing reliable and scalable geometric learning tools for BCI, neuroimaging, and related biomedical applications.

More broadly, \SPDL should be viewed as a geometric framework that connects classical geometric statistics with modern machine learning. Its importance lies in providing principled representations and learning mechanisms for structured data while maintaining compatibility with emerging AI paradigms. Recent advances in geometric deep learning have extended SPD modeling to self-supervised learning, neural dynamical systems, generative modeling, and other AI approaches, creating new opportunities for manifold-valued neuroimaging analysis.

As neuroimaging moves toward larger datasets, richer multimodal integration, and more expressive predictive models, these opportunities must be evaluated through reproducible comparisons and physiologically grounded validation. In this sense, \SPDL provides a foundation for both advancing established BCI and neuroimaging applications and developing new AI-driven tasks involving structured, manifold-valued neural data.

\section*{ACKNOWLEDGMENT}

This work was supported by grant ANR-22-PESN-0012 under the France 2030 program managed by the Agence Nationale de la Recherche (ANR), and by the DATAIA Convergence Institute under the "Programme d’Investissement d’Avenir" (ANR-17-CONV-0003), operated by Inria.

This work was also supported by the RIE2020 Industry Alignment Fund--Industry Collaboration Projects (IAF-ICP) Funding Initiative, cash and in-kind industry contributions, and the RIE2020 AME Programmatic Fund, Singapore (No.~A20G8b0102).

{
\renewcommand{\footnotesize}{\fontsize{8pt}{8.1pt}\selectfont}
\emergencystretch=2em
\sloppy
\clubpenalty=4000
\widowpenalty=4000
\bibliographystyle{IEEEtran}
\bibliography{refs}
}

\end{document}